\documentclass[10pt,twocolumn,letterpaper]{article}

\usepackage[pagenumbers]{cvpr} % To force page numbers, e.g. for an arXiv version

\usepackage{graphicx}
\usepackage{amsmath}
\usepackage{amssymb}
\usepackage{booktabs}

\usepackage{soul} 
\usepackage{booktabs}
\usepackage{multirow}
\usepackage{caption}
\usepackage{subcaption}
\usepackage{url}

\usepackage{color}
\usepackage[dvipsnames,table]{xcolor}

\usepackage{pifont}
\usepackage{colortbl}
\definecolor{Gray}{gray}{0.85}
\definecolor{SkyBlue}{rgb}{0.88,1,1}
\newcolumntype{a}{>{\columncolor{Gray}}c}

\usepackage[pagebackref,breaklinks,colorlinks]{hyperref}

\usepackage[capitalize]{cleveref}
\crefname{section}{Sec.}{Secs.}
\Crefname{section}{Section}{Sections}
\Crefname{table}{Table}{Tables}
\crefname{table}{Tab.}{Tabs.}

\def\cvprPaperID{-} % *** Enter the CVPR Paper ID here
\def\confName{CVPR}
\def\confYear{2023}

\let\OLDthebibliography\thebibliography
\renewcommand\thebibliography[1]{
  \OLDthebibliography{#1}
  \setlength{\parskip}{0pt}
  \setlength{\itemsep}{0pt plus 0.3ex}
}

\begin{document}\sloppy

% Example definitions.
% --------------------
\def\x{{\mathbf x}}
\def\L{{\cal L}}

% Title.
% ------

\title{Action-GPT: Leveraging Large-scale Language Models for Improved and Generalized Action Generation}

\author{
Sai Shashank Kalakonda\\
{\tt\small sai.shashank@research.iiit.ac.in} \\
\and Shubh Maheshwari \\ 
{\tt\small maheshwarishubh98@gmail.com} \\
\and Ravi Kiran Sarvadevabhatla \\ 
{\tt\small ravi.kiran@iiit.ac.in } \\
Centre for Visual Information Technology \\
IIIT Hyderabad, Hyderabad, INDIA 500032
}

\twocolumn[{%
\renewcommand\twocolumn[1][]{#1}%
\maketitle
\begin{center}
    \centering
    \captionsetup{type=figure}

    \includegraphics[width=\textwidth]{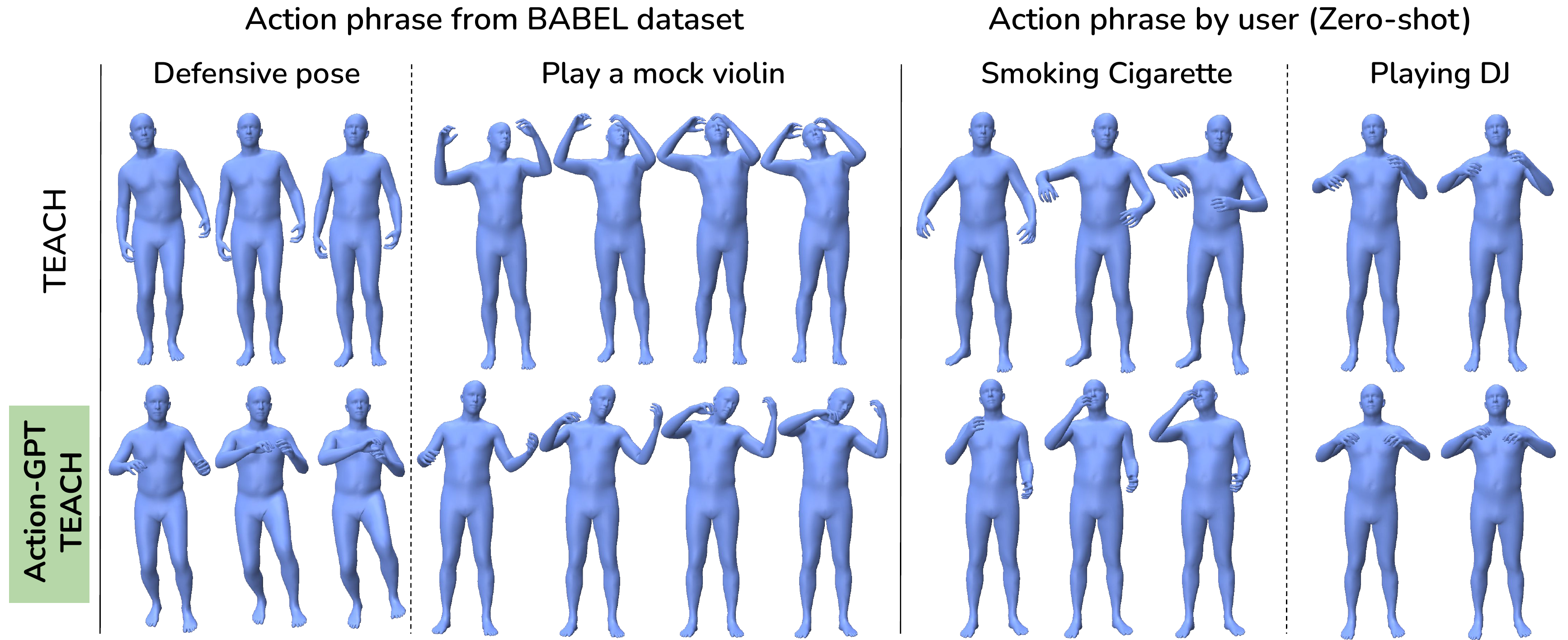}
    \captionof{figure}{Sample text conditioned action generations from a state-of-the-art model TEACH~\cite{TEACH:3DV:2022} (top row) and our large language model-based approach (Action-GPT-TEACH - bottom row). By incorporating large language models, our approach results in noticeably improved generation quality for seen and unseen categories. The conditioning action phrases for seen categories are taken from BABEL~\cite{BABEL:CVPR:2021} whereas unseen action phrases were provided by a user.}
    \label{fig:introfig}
\end{center}%
}]

\begin{abstract}

We introduce Action-GPT, a plug-and-play framework for incorporating Large Language Models (LLMs) into text-based action generation models. Action phrases in current motion capture datasets contain minimal and to-the-point information. By carefully crafting prompts for LLMs, we generate richer and fine-grained descriptions of the action. We show that utilizing these detailed descriptions instead of the original action phrases leads to better alignment of text and motion spaces. We introduce a generic approach compatible with stochastic (e.g. VAE-based) and deterministic (e.g. MotionCLIP) text-to-motion models. In addition, the approach enables multiple text descriptions to be utilized. Our experiments show (i) noticeable qualitative and quantitative improvement in the quality of synthesized motions, (ii) benefits of utilizing multiple LLM-generated descriptions, (iii) suitability of the prompt function, and (iv) zero-shot generation capabilities of the proposed approach. Code, pretrained models and sample videos will be made available at \url{https://actiongpt.github.io}.

\begin{figure*}[!h]
    \centering

    \includegraphics[width=\textwidth]{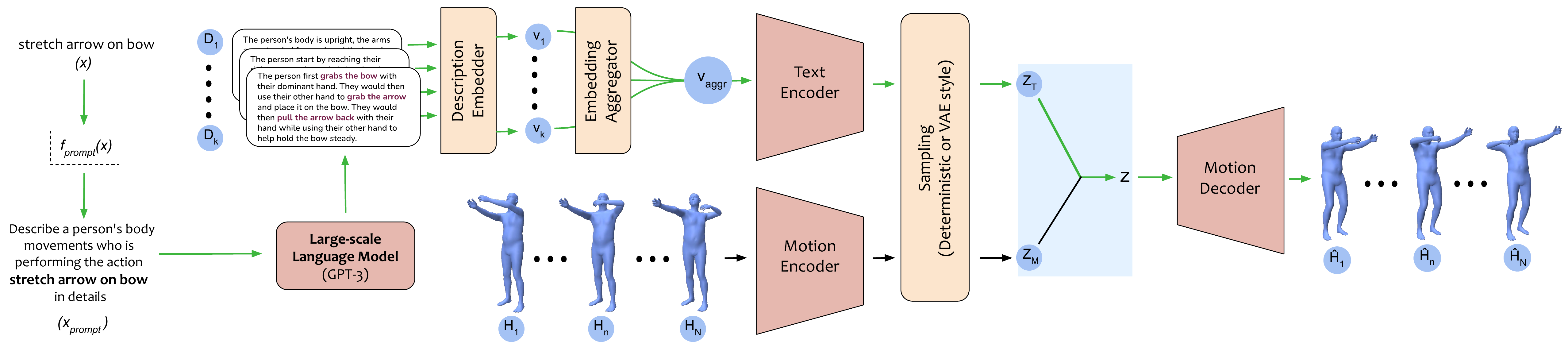}
     \caption{Action-GPT Overview: Given an action phrase ($x$), we first create a suitable prompt using an engineered prompt function $f_\text{prompt}(x)$. The result ($x_{\text{prompt}}$) is passed to a large-scale language model (GPT-3) to obtain multiple action descriptions ($D_i$) containing fine-grained body movement details.
     The corresponding deep text representations $v_i$ are obtained using Description Embedder. The aggregated version of these embeddings $v_{aggr}$ is processed by the Text Encoder. During training, the action pose sequence $H_1,...,H_N$ is processed by a Motion Encoder. The encoders are associated with a deterministic sampler (autoencoder)~\cite{tevet2022motionclip} or a VAE style generative model~\cite{petrovich22temos,TEACH:3DV:2022}.
     During training (shown with black), the latent text embedding $Z_T$ and the latent motion embedding $Z_M$ are aligned. During inference (shown in green), the sampled text embedding is provided to the Motion Decoder, which outputs the generated action sequence ${\widehat{H}}$.}
     \label{fig:overview}
\end{figure*}

\end{abstract}

\section{Introduction}
\label{sec:intro}

Human motion generation finds a vast set of applications spanning from entertainment (e.g. game and film industry) to virtual reality and robotics. There have been significant contributions on \textit{category}-conditioned human motion generations~\cite{petrovich21actor}, with some works capable of generation at scale~\cite{MUGL,DSAG}. However, the generated samples are restricted to a finite set of action categories. More recent approaches focus on \textit{text}-conditioned motion generation by constraining motion and language representations via a jointly optimized latent space~\cite{petrovich22temos,tevet2022motionclip,TEACH:3DV:2022}.

The recent development of large-scale language models (LLMs)~\cite{shoeybi2019megatronlm,gpt3} has triggered a paradigm shift in the field of Natural Language Processing (NLP). These models, pre-trained on enormous amounts of text~\cite{llmsurvey}, have demonstrated impressive generalization capabilities for challenging zero-shot setting tasks such as text generation~\cite{gpt3mix}. This exciting advance has also driven progress for various applications in computer vision~\cite{lst,cupl}, including the related task of pose-based human action recognition~\cite{lst}.

The appeal of LLM models lies in their ability to generate task-relevant text when provided a so-called \textit{prompt} - a small piece of text - as input. Motivated by this observation and the advances mentioned above, we introduce Action-GPT, an approach that utilizes the generative power of LLMs to improve the quality and generalization capabilities of action generative models. In particular, we demonstrate that our plug-and-play approach can be used to advance existing state-of-the-art motion generation architectures in a practical manner. 

Our contributions are summarized below:

\begin{itemize}
    \item To the best of our knowledge, we are the first to incorporate Large Language Models (LLMs) for text-conditioned motion generation.
    \item We introduce a carefully crafted prompt function that enables the generation of meaningful descriptions for a given action phrase.
    \item We introduce Action-GPT, a generic plug-and-play framework which is compatible with stochastic (e.g. VAE-based~\cite{TEACH:3DV:2022,petrovich22temos}) and deterministic (e.g. MotionCLIP~\cite{tevet2022motionclip}) text-to-motion models. In addition, our framework enables multiple generated text descriptions to be utilized for action generation.    
    \item Via qualitative and quantitative experiments, we demonstrate (i) noticeable improvement in the quality of synthesized motions, (ii) benefits of utilizing multiple LLM-generated descriptions, (iii) suitability of the prompt function, and (iv) zero-shot generation capabilities of the proposed approach. 
\end{itemize}

Code, pretrained models and sample videos will be made available at \url{https://actiongpt.github.io}.          

\begin{figure*}[!h]
    \centering

    \includegraphics[width=\textwidth]{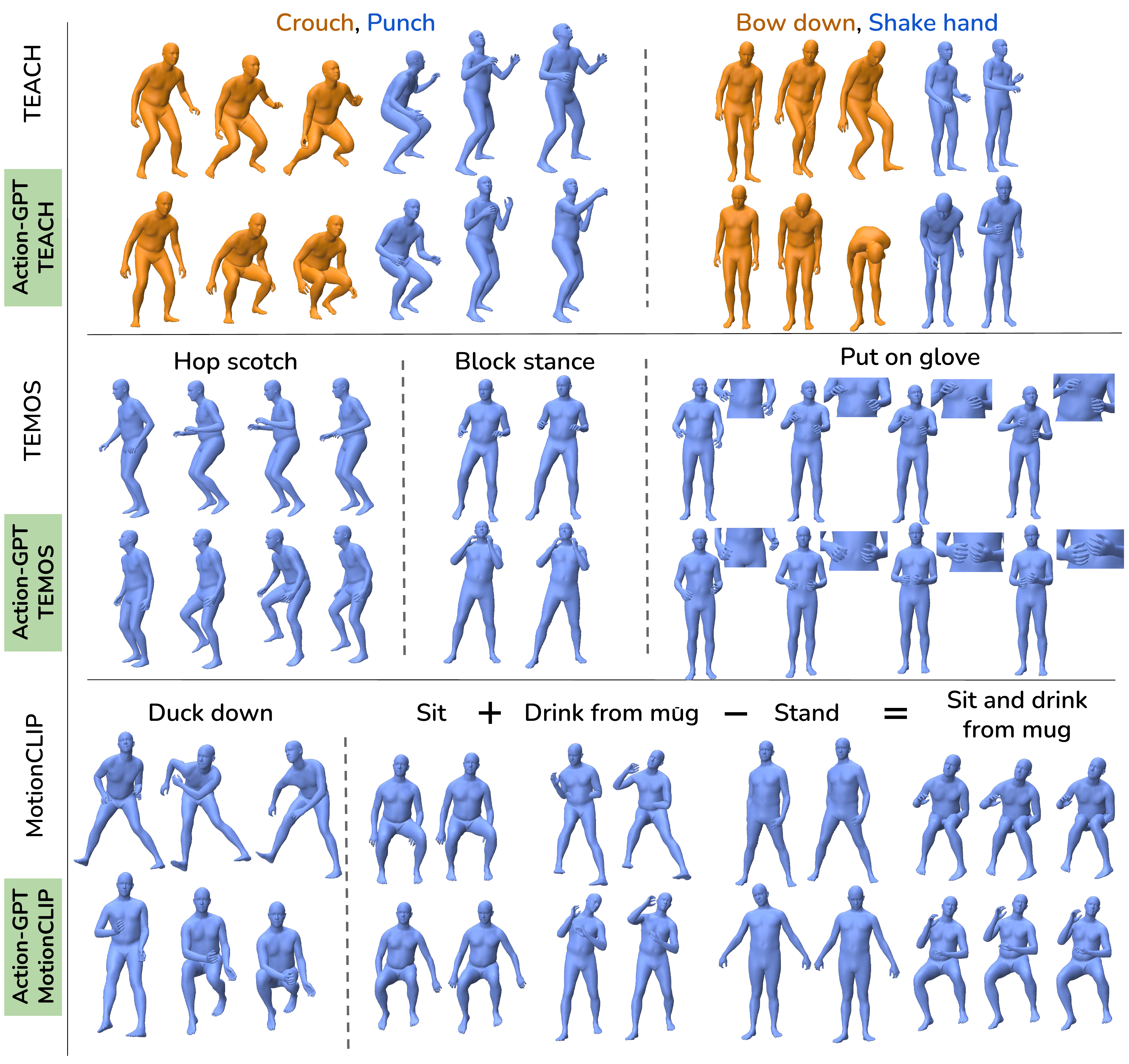}
    \caption{Visual comparison of generated motion sequences across models trained on Action-GPT framework on BABEL~\cite{BABEL:CVPR:2021} dataset. Note that the generations using Action-GPT are well-aligned with the semantic information of action phrases. The example in the bottom right row shows latent space editing. Action-GPT is better able to transfer the \textit{drink from mug} style from standing to sitting pose.}
     \label{fig:main-figure}
\end{figure*}

\section{Action-GPT}
\label{sec:actiongpt}

Our objective is to generate an actor performing the motion conditioned on the given action phrase. 
The input action phrase is a natural language text that gives a high-level description of the action. 
It is denoted as a sequence of words $x = [ w_1, w_2, ..., w_M ]$. The action is represented as a sequence of human poses ${H}= \{ H_1,  \dots, H_n, \dots, H_N  \}$ where $N$ represents the number of timesteps. The human pose $H_n \in \mathcal{R}^{|J| \times 6}$, where J is the number of joints, is the parametric SMPL~\cite{SMPL} representation wh   ich encodes the global trajectory and the parent relative joint rotation using the 6D~\cite{Zhou_2019_CVPR} rotation representation. 

Our proposed framework Action-GPT can be incorporated in an autoencoder~\cite{tevet2022motionclip} or a Variational Auto Encoder~\cite{petrovich22temos,TEACH:3DV:2022} based text-conditioned motion generation model. These motion generation models aim to generate a motion sequence conditioned on the text input by learning a joint latent space between the text and motion modalities. The key components in these models are Text Encoder $\mathcal{T}_{enc}$, Motion Encoder $\mathcal{M}_{enc}$ and Motion Decoder $\mathcal{M}_{dec}$. The two text and motion encoders encode the text sequence and motion sequence to text $Z_T$ and motion $Z_M$ latent embeddings of the same dimension, respectively. In the case of autoencoders, the latent embeddings are obtained in a deterministic fashion, whereas in Variational Auto Encoders, the latent embeddings are sampled from the Gaussian distribution $\mathcal{N}(\mu,\Sigma)$, where $(\mu,\Sigma)$ are the outputs of the encoder. The motion decoder, on the other hand, uses the latent embedding $Z$ as input to generate a sequence of motion poses ${\widehat{H}}= \{ \widehat{H}_1,  \dots, \widehat{H}_n, \dots, \widehat{H}_N  \}$

Fig.~\ref{fig:overview} provides an overview of our approach to incorporate LLM (GPT-3 in our case) into the text-conditioned motion generation models. In contrast to training directly using the action phrase $x$ from the dataset, our framework uses carefully crafted GPT-3 generated text descriptions $D_i$, which provide low-level details about the movement of individual body parts. The proposed framework consists of three steps (1) Constructing a prompt function $f_{prompt}$, (2) Aggregating multiple GPT-3 generated text descriptions $D_i$, and finally, (3) utilizing the GPT-3 generated text descriptions $D_i$ in T2M models.

\subsection{Prompt strategy}\label{sec:prompt-strategy}

\begin{figure*}[!h]
    \centering

    \includegraphics[width=\textwidth]{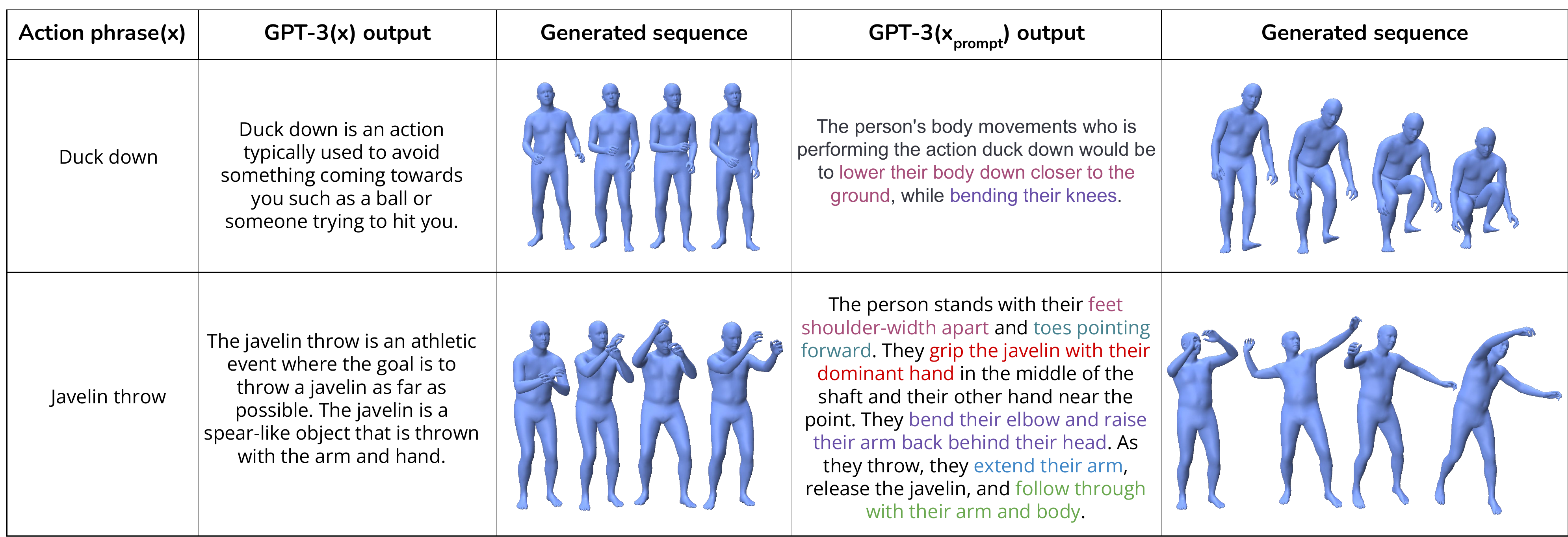}
     \caption{This figure highlights the importance of the prompt function. Observe that directly feeding the action phrase text ($x$) to GPT-3 results in poor-quality generations. In contrast, the fine-grained body movement details in the prompt-based text enable higher fidelity generations (last column). Note that the coloured text descriptions correspond to different body movement details.}
     \label{fig:prompt-importance}
\end{figure*}

\begin{figure*}[!h]
    \centering
    \includegraphics[width=\linewidth]{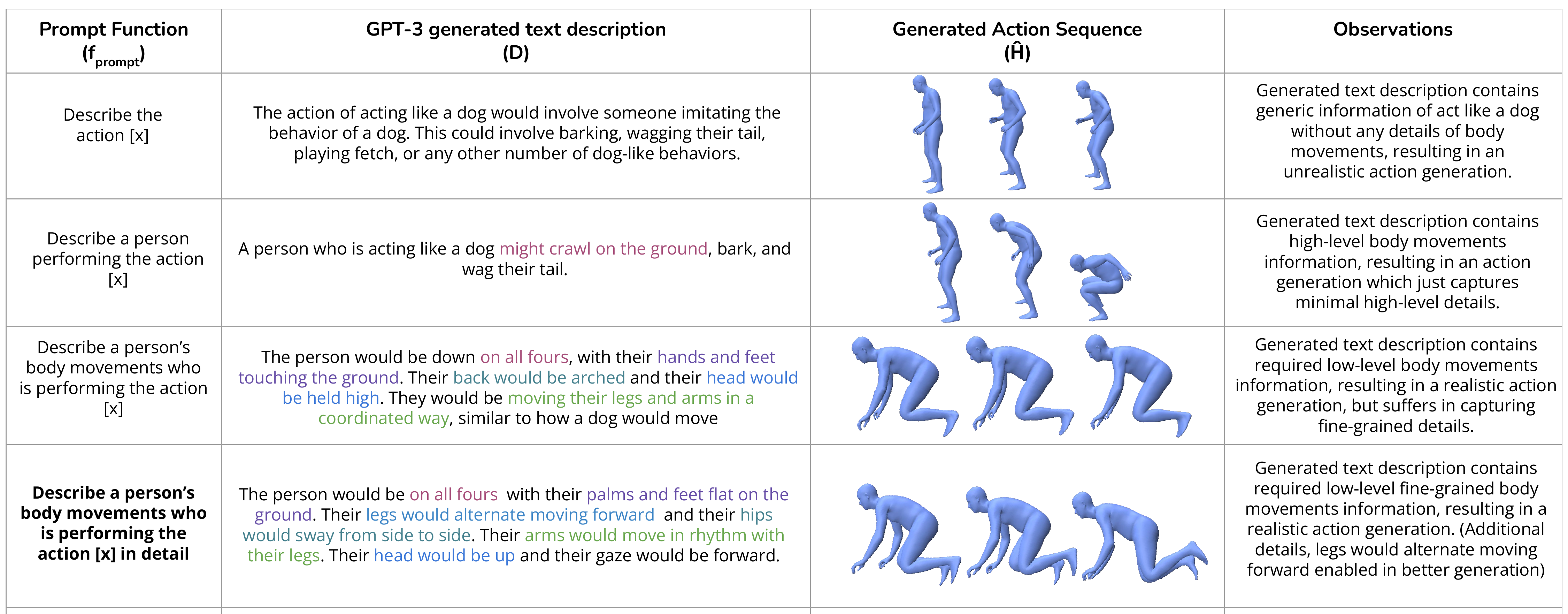}
     \caption{The table showcases the descriptions generated by GPT-3 (D), generated action sequences         (${\widehat{H}}$) for the action phrase (x =) \textit{act like a dog} using different prompt strategies along with the observations (right most column). Notice that our prompt function (bottom row) generates the highest amount of required body movement descriptions, generating the most realistic action sequence. Note that the coloured text descriptions correspond to the body movement details. }
     \label{tab:prompts}
\end{figure*}

For a given action phrase $x$, we generate low-level body movement details using GPT-3~\cite{gpt3}. GPT-3 is an autoregressive transformer model which generates human-like textual descriptions relevant to the small amount of input text provided. However, directly providing the action phrase as input to GPT-3 fails to output text containing the desired detail body movement information and leads to unrealistic motion generations (see Fig.~\ref{fig:prompt-importance}). This necessitates the need for a suitable prompt function~\cite{llmsurvey}. After multiple empirical trials, we determine the following prompting function $f_{prompt}$: \texttt{Describe a person's body movements who is performing the action [x] in detail}. Specifically, adding \texttt{Describe a person's} to the prompt restricts the description from generic information to character movement. The phrase \texttt{body movements} forces GPT-3 to explain the motion of individual body parts. Lastly, \texttt{in detail} forces the descriptions to provide low-level details. Fig.~\ref{tab:prompts} showcases the importance of each component of our prompt function. We provide GPT-3 generated text description $(D)$ and corresponding generated action sequence ${\widehat{H}}$ for the action phrase \texttt{act like a dog} along with the observations in the rightmost column.

\subsection{Aggregating multiple descriptions}

\begin{figure*}[!h]
    \centering

    \includegraphics[width=\textwidth]{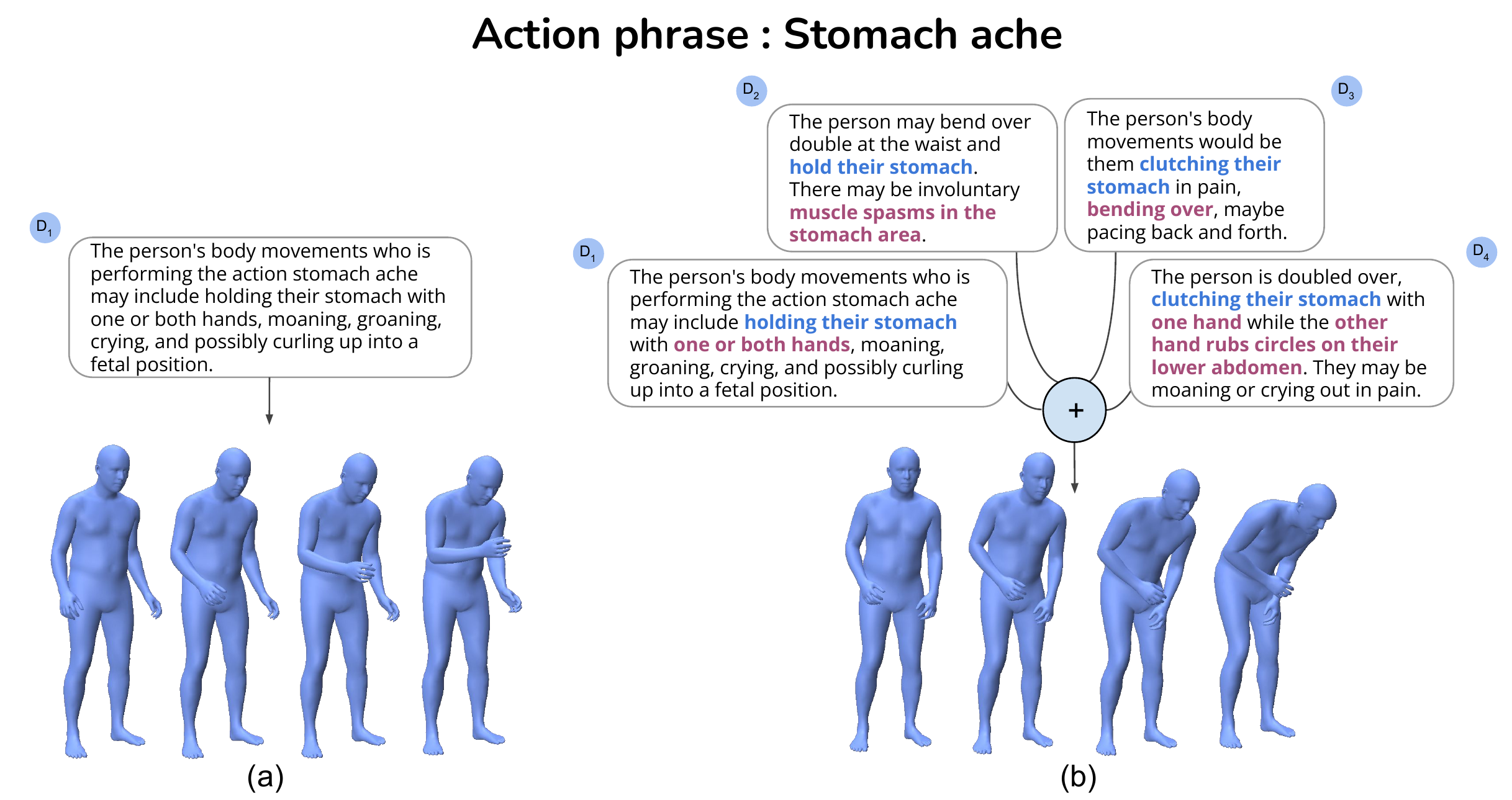}
     \caption{This figure highlights the importance of using multiple GPT-3 generated descriptions ($D_1,...,D_k$), $k=4$ for each action phrase in Action-GPT framework for TEACH~\cite{TEACH:3DV:2022}. Notice the visibly improved generation quality when multiple prompted descriptions are used (right column). Body movement text common across descriptions is highlighted in blue. Movements unique to each description are highlighted in pink.}
     \label{fig:k-importance}
\end{figure*}

Given an action prompt $x_{prompt}$, GPT-3 is capable of generating multiple textual descriptions $D_1,\dots,D_k$ describing the action-specific information. The randomly generated $k$ descriptions contain common and description-specific text segments, which enhance the overall richness of action description (see Fig.~\ref{fig:k-importance}). Therefore, we utilize multiple descriptions as part of the text-processing pipeline. The GPT-3 generated $k$ text descriptions $D_{1}, \dots, D_{k}$ are passed through a Description Embedder $D_{emb}$ to obtain corresponding description embeddings $v_{1}, \dots, v_{k}$. These $k$ description embeddings are aggregated into a single embedding $v_{aggr}$ using an Embedding Aggregator $E_{aggr}$. We consider average operation as our Embedding Aggregator unless stated otherwise.

\subsection{Utilizing GPT-3 generated text descriptions in T2M models}

The text encoder $\mathcal{T}_{enc}$ inputs the aggregated embedding $v_{aggr}$, and the outputs are sampled to generate text latent embeddings $Z_T$. In a similar fashion, motion encoder $\mathcal{M}_{enc}$ inputs the sequence of motion poses ${H}$ = \{${H}_1,\dots,{H}_N$\}, where ${H}_n \in \mathcal{R}^{|J| \times 6}$ and samples motion latent embeddings $Z_M$. In a deterministic approach~\cite{tevet2022motionclip}, the latent embeddings are generated directly as outputs of the encoder, whereas in a VAE-based approach~\cite{petrovich22temos,TEACH:3DV:2022}, the encoders generate distribution parameters $\mu$ and $\Sigma$. The latent embeddings are then sampled from the Gaussian distributions $N(\mu$, $\Sigma)$. On the other hand, the generated text and motion embeddings are provided to the motion decoder, which generates the human motion sequence ${\widehat{H}}$ =  \{${\widehat{H}}_1,\dots,{\widehat{H}}_N$\}, where ${\widehat{H}}_i \in \mathcal{R}^{|J| \times 6}$ which is passed through forward kinematics to generate corresponding 3D mesh sequence.

\section{Experiments}

% \st{below info from BABEL official paper}
\textbf{BABEL}~\cite{BABEL:CVPR:2021} is a large dataset with language labels describing the actions being performed in motion capture sequences. It contains about 43 hours of mocap sequences, comprising over 65k textual labels which belong to over 250 unique action categories. We primarily focus our results on BABEL, considering its vast and diverse set of motion sequences assigned to short text sequences, which contain an average of 3-4 words. The action phrases of the BABEL dataset are to the point and precise about the action information without any additional details about the actor.

\subsection{Models}

We demonstrate our framework on state-of-the-art text conditioned motion generation models -- TEMOS~\cite{BABEL:CVPR:2021}, MotionCLIP~\cite{tevet2022motionclip} and TEACH~\cite{TEACH:3DV:2022}. Since TEACH is an extension of TEMOS and uses only pairs of motion data of BABEL, we also demonstrate our results on TEMOS by retraining it on all single action data segments of BABEL. We train these three models as per our framework using their publicly available codes and will call them Action-GPT-[model] further. 

\noindent \textbf{Action-GPT-MotionCLIP:} Similar to MotionCLIP~\cite{tevet2022motionclip}, we use CLIP's pretrained text encoder~\cite{clip} as our description embedder, but we inputs all the $k$ text descriptions and generate the aggregated vector representation $v_{aggr}$ using the embedding aggregator. Note that similar to MotionCLIP, we use CLIP-ViT-B/32 frozen model. We follow the same motion auto-encoder setup as that of MotionCLIP. There is no additional text encoder and sampling process as the constructed $v_{aggr}$ itself is used as the text embedding $Z_T$, and the output of the motion encoder is used as the motion embedding $Z_M $.     

\noindent \textbf{Action-GPT-TEMOS:} Instead of providing action phrase text, we pass multiple textual descriptions extracted from GPT-3 to DistilBERT~\cite{distilbert} to obtain description embedding $v_i \in R^{n_{i} \times e}$, where $n_i$ is the number of words in description $D_i$ and $e$ is the DistilBERT embedding dimension, thus $v_{aggr} \in R^{G \times e}$ where $G=max(n_i)$. Note that similar to TEMOS, we use pre-trained DistilBERT and freeze its weights during training. The text encoder, motion encoder, and motion decoder used is the same as that of TEMOS. The sampled text embedding $Z_T$ and motion embedding $Z_M$ are both $\in R^{d}$ where $d$ is the dimension of latent space.

\noindent \textbf{Action-GPT-TEACH:} Since TEACH~\cite{TEACH:3DV:2022} is an extension of TEMOS, the process of generating description embeddings $v_i$ is the same as that of in Action-GPT-TEMOS.  The text encoder, motion encoder, and motion decoder are used the same as that of TEACH.  As TEACH is trained on the pairs of action data, the training iteration consists of two forward passes where an action phrase and its corresponding motion sequence are provided as input in each pass. In addition, a set of the last few frames of generated motion in the first pass are also provided as input in the second pass. In both passes, our framework uses the generated description embeddings corresponding to the input action phrases.

Detailed diagrams illustrating the differences between original architectures and their LLM-based variants along with the additional details regarding training and testing can be found in the appendix~\ref{sec:model-details}.

\subsection{Implementation details}
\label{sec:gpt3-config}

We access GPT-3 via OpenAI API Beta Access program. Unless stated otherwise, we use the largest GPT-3 model available, \texttt{davinci-002}. The Action-GPT prompt strategy consumes a maximum of 140 tokens together for prompt and generation. We use the completions API endpoint with the parameters temperature and top-p set to 0.5 and 1, ensuring we have well-defined diverse descriptions. All the other parameters are set to default. We conduct all our experiments on cluster machines with Intel Xenon E5 2640 v4 and Nvidia GeForce GTX Ti 12GB GPUs with Ubuntu 16.04 OS.

\subsection{Quantitative analysis}

We follow the metrics employed in TEACH~\cite{TEACH:3DV:2022} for quantitative evaluation, namely Average Positional Error (APE) and Average Variational Error (AVE), measured on the root joint and the rest of the body joints separately. Mean local correspond to the joint position in the local coordinate system (with respect to the root), whereas mean global corresponds to the joint position in the global coordinate system. The APE and AVE for a joint are the averages of the L2 distances between the generated and ground truth joint positions and variances, respectively -- refer to appendix~\ref{sec:metrics-details} for details. Tab.~\ref{tab:comparisions} summarizes the results of using our framework in comparison with the default setup for each model. Incorporating detailed descriptions using GPT-3 shows an improvement over all the APE (except for MotionCLIP) and AVE metrics. The metrics of root joints for MotionCLIP are empty since it generates only local pose without any locomotion. 

\subsection{Qualitative analysis}

In Fig.~\ref{fig:main-figure}, we provide qualitative comparisons of the model generations. We observe that the generations from our framework are more realistic and well-aligned with the semantic information of the action phrases compared to the default approach. The generations are able to capture the low-level fine-grained details of the action suggested by the original text phrase input. Please refer to appendix for video examples of zero-shot generations, comparisons with baselines, and examples showing diversity in generated sequences.

\begin{table*}[!t]
\centering
\resizebox{0.9\linewidth}{!}
{
\begin{tabular}{ c|l|cccc|cccc}
% \toprule
&  & \multicolumn{4}{c}{\textbf{Average Positional Error ↓ }}                                                    & \multicolumn{4}{c}{\textbf{Average Variance Error ↓ }} \\
\midrule
Model & Method  & root joint & global traj & mean local & mean global & root joint & global traj & mean local & mean global \\

\midrule

\multirow[c]{2}{*}[-0.5em]{\rule{0pt}{2ex} MotionCLIP}
& Default & - & - & 0.556 & 0.541 & - & - & 0.056 & 0.02  \\
& Action-GPT & - & - & 0.590 & 0.571 & - & - & \textbf{0.042} & \textbf{0.019} \\
% \cline{1-10} 
\midrule
\multirow[c]{2}{*}[-0.5em]{\rule{0pt}{2ex} TEMOS}
& Default & 0.597 & 0.574 & 0.162 & 0.644 & 0.113 & 0.112 & 0.010 & 0.122  \\
& Action-GPT & \textbf{0.561} & \textbf{0.540} & \textbf{0.151} & \textbf{0.605} & \textbf{0.101} & \textbf{0.100} & \textbf{0.010} & \textbf{0.109} \\
% \cline{1-10} 
\midrule
\multirow[c]{2}{*}[-0.5em]{\rule{0pt}{2ex} TEACH}
& Default & 0.674 & 0.654 & 0.159 & 0.717 & 0.222 & 0.220 & 0.014 & 0.234  \\
& Action-GPT & \textbf{0.606} & \textbf{0.586} & \textbf{0.159} & \textbf{0.650} & \textbf{0.204} & \textbf{0.202} & \textbf{0.014} & \textbf{0.216} \\
% \cline{1-10} 
\bottomrule
\end{tabular}
}
\captionof{table}{Quantitative evaluation on the BABEL test set} 
\label{tab:comparisions}
\end{table*}

\begin{table*}[!t]
\centering
\resizebox{0.9\linewidth}{!}
{
\begin{tabular}{ c|l|cccc|cccc}
% \toprule
&  & \multicolumn{4}{c}{\textbf{Average Positional Error ↓ }}                                                    & \multicolumn{4}{c}{\textbf{Average Variance Error ↓ }} \\
\midrule
Architectural Component & Ablation Details  & root joint & global traj & mean local & mean global & root joint & global traj & mean local & mean global \\

\midrule

% \multirow[c]{5}{*}[-2.5em]{\rule{0pt}{2ex}Architectural} &
\multirow[c]{2}{*}[-0.5em]{\rule{0pt}{2ex} number of generated descriptions ($K$)}
& $k=1$ & 0.655 & 0.635 & 0.159 & 0.698 & 0.216 & 0.214 & 0.015 & 0.228  \\
& $k=2$ & 0.637 & 0.617 & 0.158 & 0.680 & 0.211 & 0.209 & 0.015 & 0.223 \\
% & $K=4$ & 0.606 & 0.586 & 0.158 & 0.65 & 0.204 & 0.202 & 0.014 & 0.216 \\
& $k=8$ & 0.632 & 0.613 & \textbf{0.157} & 0.674 & 0.212 & 0.210 & 0.015 & 0.224 \\
% \cline{1-10} 
\midrule
GPT-3 Capacity & \texttt{curie} & 0.642 & 0.622 & 0.159 & 0.680 & 0.216 & 0.214 & 0.015 & 0.228 \\
% \hline
\midrule

\textbf{Ours} ($\mathbf{k=4}$) & \texttt{davinci} & \textbf{0.606} & \textbf{0.586} & 0.158 & \textbf{0.650} & \textbf{0.204} & \textbf{0.202} & \textbf{0.014} & \textbf{0.216} \\ 
\bottomrule
\end{tabular}
}
\captionof{table}{Performance scores for ablative variants.} 
\label{tab:ablations}
\end{table*}

\subsection{Ablations}

We perform an ablation study to understand the underlying effects of the Action-GPT framework. All of the ablation experiments are carried out on the Action-GPT-TEACH model unless stated otherwise, as it is capable of handling a series of action phrases as input.

\noindent \textbf{Number of GPT-3 Text Sequences}: We analyzed the influence of number of generated descriptions in Action-GPT-TEACH framework by varying $k$ in $1,2,4$ and $8$. We observed that for all the values of $k$, Action-GPT-TEACH performs better than the default TEACH and the best results are obtained for $k=4$ (see Tab.~\ref{tab:ablations}). Increasing the value of $k$ up to a certain value improves performance. However,  aggregating too many descriptions can lead to the injection of excessive noise, which dilutes the presence of text related to body movement details. 

\noindent \textbf{Language Model Capacity}: Open AI provides GPT-3 in different model capacities, \texttt{davinci} being the largest. We analyzed the influence of \texttt{curie}, the second largest GPT-3 model, on the motion sequence generations of the Action-GPT-TEACH $(k=4)$ framework. Results show that having a larger model capacity helps in generating more realistic motion sequences, as the generated text descriptions provide much relevant and detailed information as required.

\section{Discussion and Conclusion}

The key to good quality and generalizable text-conditioned action generation models lies in improving the alignment between the text and motion representations. Through our Action-GPT framework, we show that such alignment can be achieved efficiently by employing Large Language Models whose operation is guided by a judiciously crafted prompt function. Sentences in GPT-generated descriptions contain procedural text which corresponds to sub-actions. During training, the regularity and frequency of such procedural text likely enable better alignment with corresponding motion sequence counterparts. We also hypothesize that the diversity of procedural sentences in the descriptions enables better compositionality for unseen (zero-shot) generation settings.    

The plug-and-play nature of our approach is practical for adoption within state-of-the-art text-conditioned action generation models. Our experimental results demonstrate the generalization capabilities and action fidelity improvement for multiple adopted models, qualitatively and quantitatively. In addition, we also highlight the role of various prompt function components and the benefit of utilizing multiple prompts for improved generation quality.

\bibliographystyle{IEEEbib}
\bibliography{main}

\clearpage

\noindent \textbf{\LARGE Appendix}
\section{Models}

\label{sec:model-details}We demonstrate our framework on state-of-the-art text conditioned motion generation models -- TEMOS~\cite{BABEL:CVPR:2021}, MotionCLIP~\cite{tevet2022motionclip} and TEACH~\cite{TEACH:3DV:2022}, all trained on BABEL~\cite{BABEL:CVPR:2021}. We name their LLM (i.e. GPT here) based variants as Action-GPT-[model]. 

\begin{figure*}[!h]
    \centering
    \includegraphics[width=\textwidth]{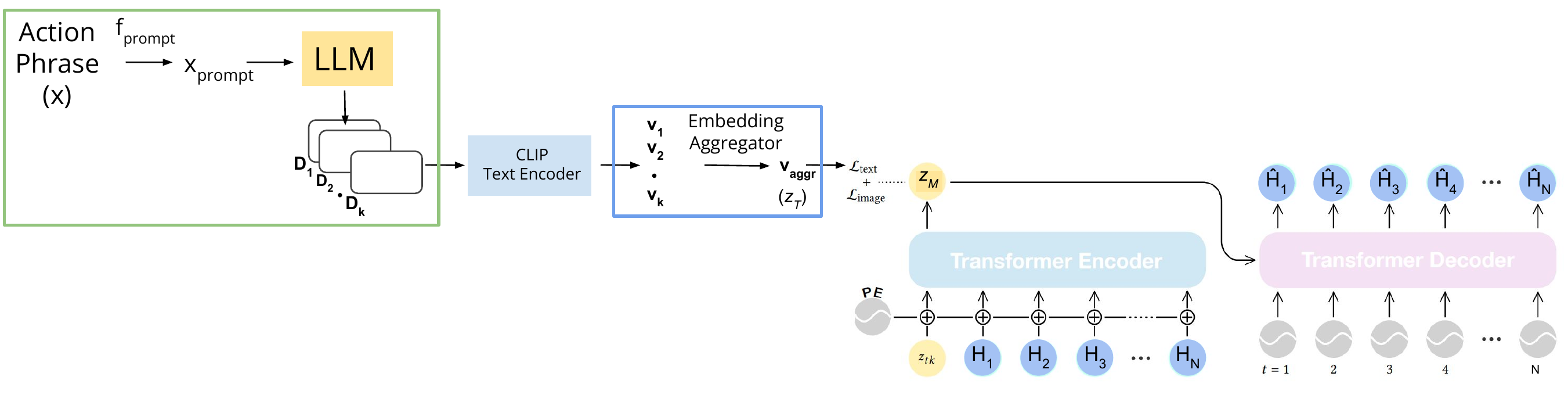}
     \caption{Action-GPT-MotionCLIP Overview: We extend MotionCLIP~\cite{tevet2022motionclip} by incorporating LLM (i.e. GPT-3). The box highlighted in green showcases the generation of $k$ text descriptions $D_i$ as the output of LLM on input $x_{prompt}$, which is constructed using the prompt function $f_{prompt}$ and action phrase $x$. The box highlighted in blue showcases the aggregation of description embeddings $v_i$, outputs of the CLIP text encoder. The aggregated embedding $v_{aggr}$ is considered as the latent text embedding $Z_T$, on which the text loss $\mathcal{L}_{text}$ is computed. All the other components apart from the highlighted boxes represent the original architecture of MotionCLIP~\cite{tevet2022motionclip}.}
     \label{fig:actgpt_motionclip}
\end{figure*}

\noindent \textbf{Action-GPT-MotionCLIP:} In MotionCLIP~\cite{tevet2022motionclip}, for a given action phrase $x$, the CLIP text embedding of the phrase $x$ is considered as its corresponding latent text embedding $Z_T$, whereas in our Action-GPT framework the aggregated vector embedding $v_{aggr} \in R^{c}$, where $c$ is the CLIP text embedding dimesnion, constructed for the action phrase $x$ is considered as its latent text embedding $Z_T$. In detail, we first construct $x_{prompt}$ using the action phrase $x$ and prompt function $f_{prompt}$. The $x_{prompt}$ is then input to LLM (i.e. GPT-3) to generate $k$ textual descriptions $D_{i}$. Using the CLIP text encoder, we then construct $k$ corresponding CLIP text embeddings $v_i$. These $k$ CLIP text embeddings $v_i$ are then aggregated into a single embedding $v_{aggr}$ using an Embedding aggregator (average operation here). The constructed $v_{aggr}$ is the corresponding latent text embedding $Z_T$ for the action phrase $x$. (see Fig.~\ref{fig:actgpt_motionclip})

We trained this model using the same training configurations as mentioned in MotionCLIP~\cite{tevet2022motionclip} and provided the results on the test split. We generated the metrics for the baseline MotionCLIP~\cite{tevet2022motionclip} using the pre-trained model provided.

\begin{figure}[!h]
    \centering
    \includegraphics[width=\linewidth]{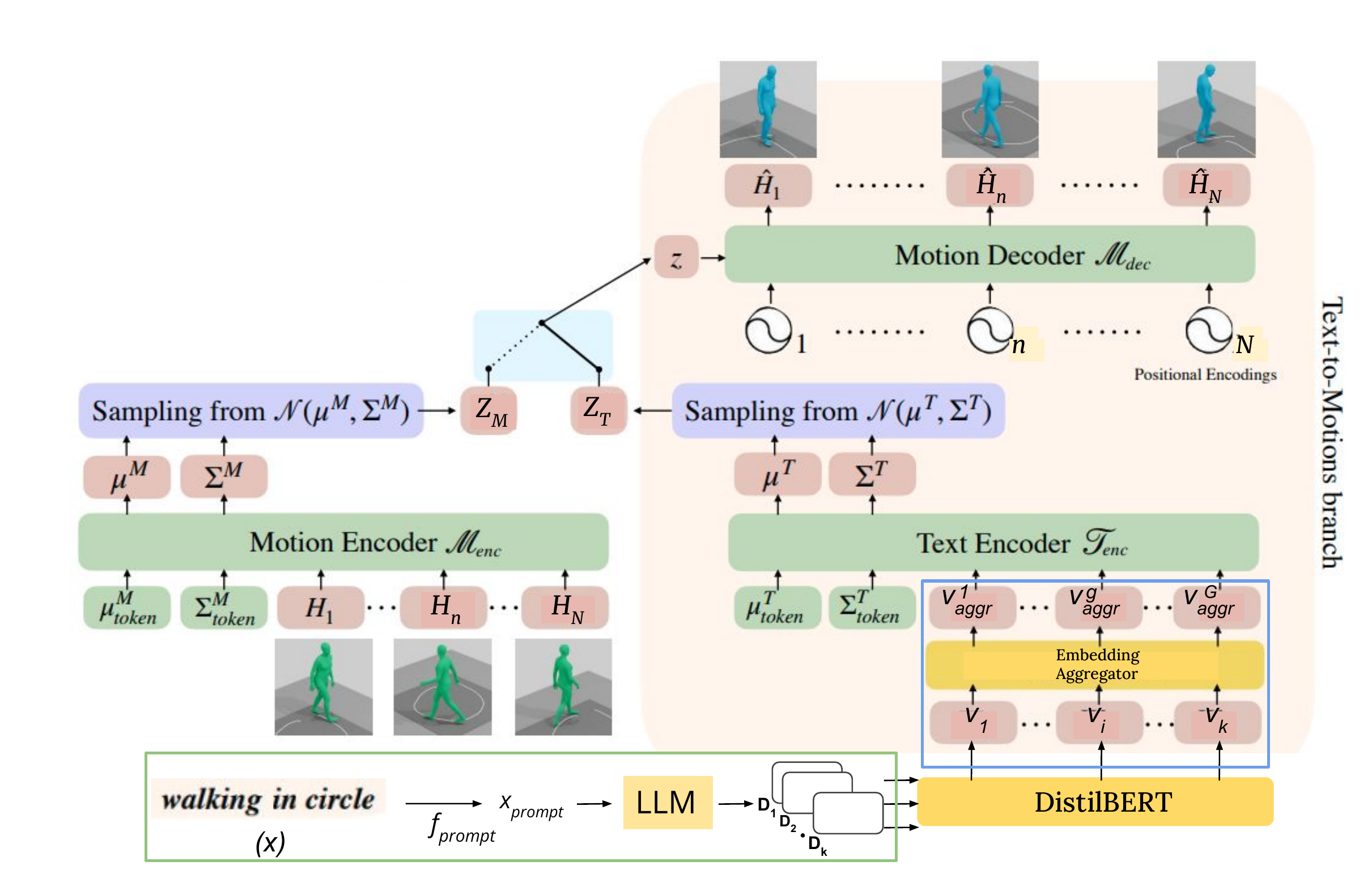}
     \caption{Action-GPT-TEMOS Overview: We extend TEMOS~\cite{petrovich22temos} by incorporating LLM (i.e. GPT-3). The box highlighted in green showcases the generation of $k$ text descriptions $D_i$ using LLM on input $x_{prompt}$, which is constructed using the prompt function $f_{prompt}$ and action phrase $x$. The $k$ text descriptions $D_i$ are input to DistilBERT to generate corresponding description embeddings $v_i$. The box highlighted in blue showcases the aggregation of description embeddings $v_i$ to $v_{aggr}^{1:G}$ using an embedding aggregator. All the other components apart from the highlighted boxes represent the original architecture of TEMOS~\cite{petrovich22temos}.} 
     \label{fig:actgpt_temos}
\end{figure}

\noindent \textbf{Action-GPT-TEMOS:} Instead of providing the action phrase $x$, directly to DistilBERT~\cite{distilbert} as that in TEMOS~\cite{petrovich22temos}, we first construct $x_{prompt}$ and generate $k$ textual descriptions $D_i$ using LLM (i.e. GPT-3). These $k$ textual descriptions are then input to DistilBERT~\cite{distilbert} to obtain $k$ corresponding description embeddings $v_i \in R^{n_{i} \times e}$, where $n_i$ is the number of words in description $D_i$ and $e$ is the DistilBERT embedding dimension. The $k$ description embeddings are aggregated to a single embedding $v_{aggr} \in R^{G \times e}$, where $G = max(n_i)$ using the average operation. This aggregated embedding $v_{aggr}$ is then used as input to Text Encoder along with the text distribution tokens $\mu_{token}^T$ and $\sum_{token}^T$. The later training and inference process is carried out the same as that of in TEMOS~\cite{petrovich22temos}. (see Fig.~\ref{fig:actgpt_temos})

We trained this model on a 4 GPU setup with a batch size of 4 on each GPU, keeping rest all the parameters same as provided in TEMOS~\cite{petrovich22temos}. We trained the baseline model of TEMOS~\cite{petrovich22temos} and its LLM-based variant on BABEL~\cite{BABEL:CVPR:2021} using the single action data segments provided by TEACH~\cite{TEACH:3DV:2022} and generated the metrics on the corresponding test set.

\begin{figure}[!h]
    \centering
     \includegraphics[width=\linewidth]{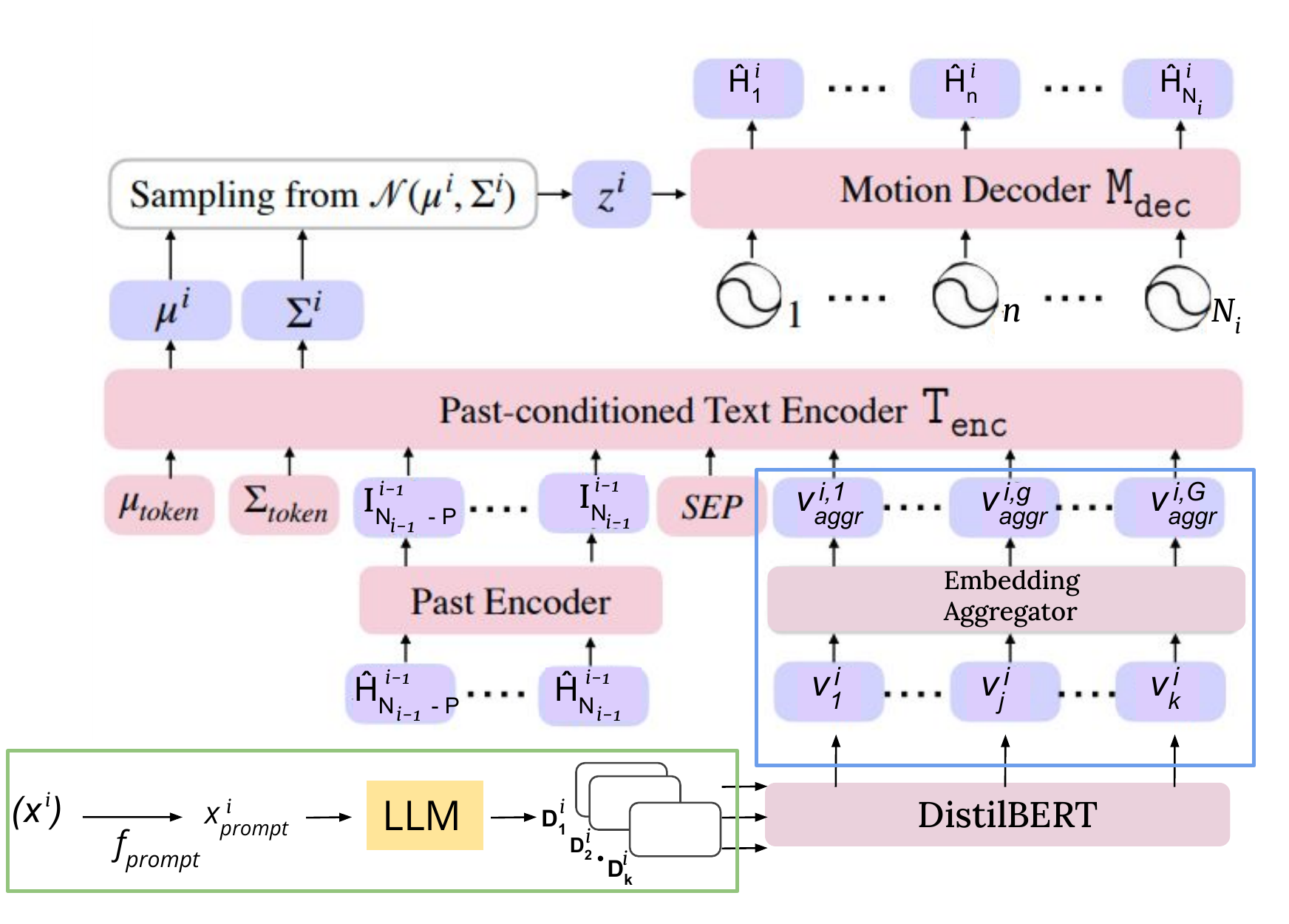}
     \caption{Action-GPT-TEACH Overview: We extend TEACH~\cite{TEACH:3DV:2022} by incorporating LLM (i.e. GPT-3). As TEACH can generate an action sequence for a series of action phrases, we use $x^i$ to denote the $i^{th}$ phrase. The box highlighted in green showcases the $k$ generated text descriptions $D_j^i$ using LLM on input $x_{prompt}^i$, which is constructed using the prompt function $f_{prompt}$ and action phrase $x^i$. The $k$ text descriptions $D_j^i$ are input to DistilBERT to generate corresponding sentence embeddings $v_j^i$, which are aggregated into a single embedding $v_{aggr}^{i,1:G}$ using an embedding aggregator. All the other components apart from the highlighted boxes represent the original architecture of TEACH~\cite{TEACH:3DV:2022}.}
     \label{fig:actgpt_teach}
\end{figure}

\noindent \textbf{Action-GPT-TEACH:} Since TEACH~\cite{TEACH:3DV:2022} is an extension of TEMOS~\cite{petrovich22temos} the process of generating aggregated description embedding $v_{aggr}$ is same as that of Action-GPT-TEMOS. In addition, TEACH can generate an action sequence for a series of action phrases as input $x = \{x^1,\dots,x^i,\dots,x^s\}$. So, we compute $v_{aggr}$ $s$ number of times, once for each phrase $x^i$. In detail, for an action phrase $x^i$ we generate $k$ text descriptions $D_1^i,..,D_j^i,..,D_k^i$. The $k$ text descriptions are input to DistilBERT to generate corresponding description embeddings $v_1^i,..,v_j^i,..,v_k^i$, where $v_j^i \in R^{n_{j}^i \times e}$, where $n_j^i$ is the number of words in description $D_j^i$ and $e$ is the DistilBERT embedding dimension. The $k$ description embeddings $v_j^i$ are aggregated to a single embedding $v_{aggr}^{i} \in R^{G \times e}$, where $G = max(n_j^i)$ using the average operation. The aggregated embedding $v_{aggr}^{i}$ is input to Past-conditioned Text Encoder $T_{enc}$ along with the learnable tokens {$\mu_{token}, \sum_{token},$ SEP token} and ${{I}}^{i-1}_{N_{i-1}-P:N_{i-1}}$, the motion features generated using Past Encoder, corresponding to the last $P$ frames of previous generated action sequence ${\widehat{H}}^{i-1}_{N_{i-1}-P:N_{i-1}}$. The later training and inference process followed is the same as that of TEACH.(see Fig.~\ref{fig:actgpt_teach})

Similar to Action-GPT-TEMOS, we trained this model on a 4 GPU setup with a batch size of 4 on each GPU, keeping rest all the parameters same as provided in TEACH~\cite{TEACH:3DV:2022}. We generated the metrics for the baseline TEACH~\cite{TEACH:3DV:2022} using the pre-trained model provided.

\section{Diverse Generations}
Our Action-GPT framework can generate diverse action sequences for a given action phrase $x$, utilizing the capability of LLMs to generate diverse text descriptions for a given prompt. We generate multiple text descriptions $D_i$ for an action phrase $x$, and using them as input, multiple action sequences ${\widehat{H}}_i$ are generated.

\section{Zero-Shot Generations}
Our approach enables the generation of action sequences ${\widehat{H}}$ for unseen action phrases (zero-shot). The intuition behind this is that our Action-GPT framework uses low-level detailed body movements textual information to align text and motion spaces instead of using the action phrase directly. Hence the action phrase might be unseen,   but the low-level body movement details in the generated corresponding text descriptions $D_i$ will not be completely unseen.

Results corresponding to comparison of the baseline models to their LLM-based variants, diverse and zero-shot generations can be found at \url{https://actiongpt.github.io}. 

\begin{figure*}[t]
    \centering
    \includegraphics[width=0.9\textwidth]{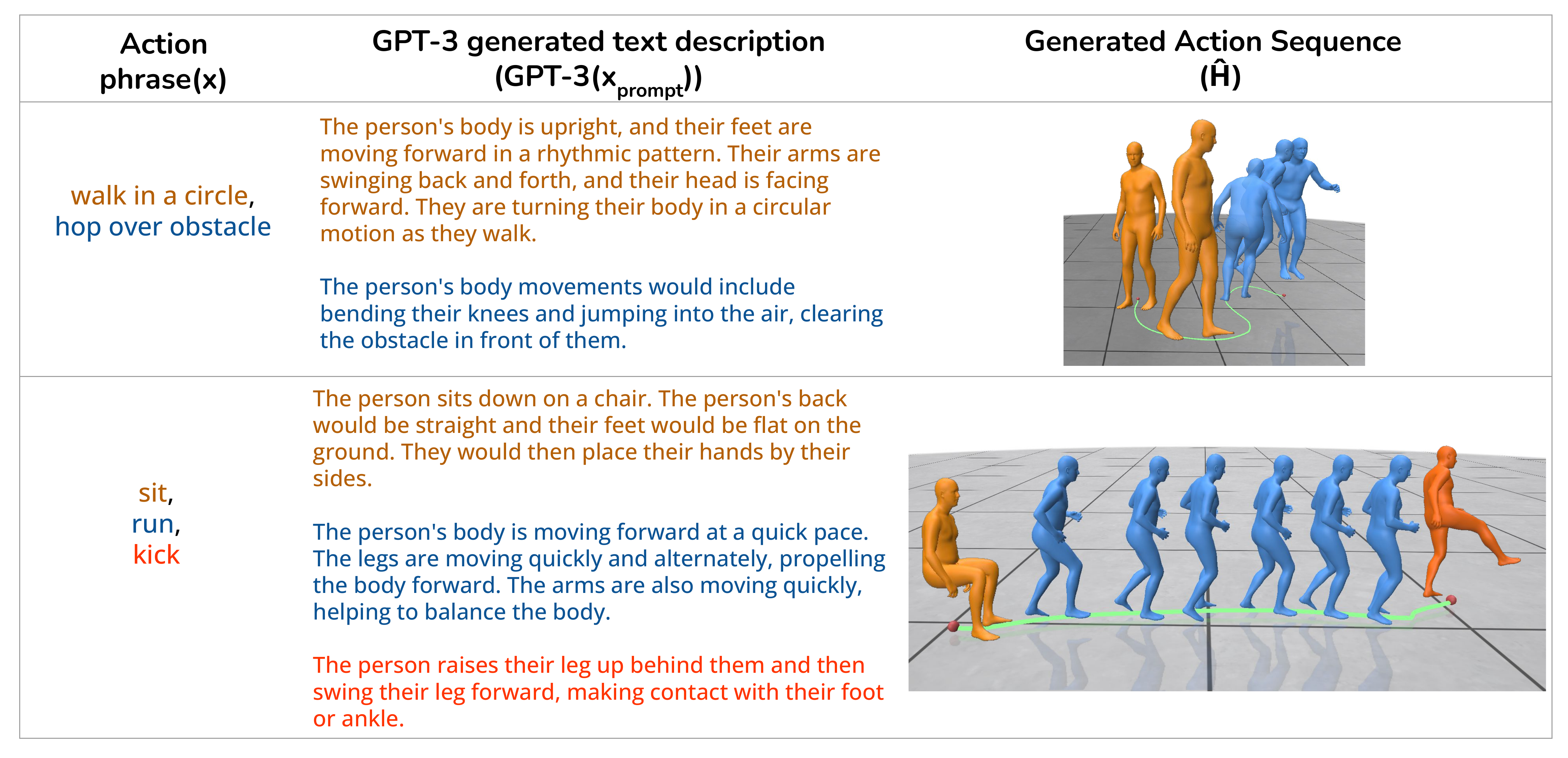}
     \caption{Actions with locomotion generated using Action-GPT-TEACH. The color of the text (column 2) represents the detailed text generated by GPT-3 and the same color of the mesh represents the pose sequence generated for the sub-action. The green curve shows the trajectory. The red points show the starting and end points of the motion. Action-GPT is able to generate diverse examples involving locomotion such as walk in a circle, run, hop over obstacle and kick.}
     \label{fig:locomotion}
\end{figure*}

\section{Metrics}
\label{sec:metrics-details}We show results using the generative quality metrics followed by \cite{petrovich22temos,TEACH:3DV:2022}, namely Average Positional Error (APE) and Average Variational Error (AVE). For all the metrics, the smaller the score, the better the generative quality.

\noindent \textbf{Average Positional Error (APE)} : For a joint j, APE is calculated as the average of the L2 distances between the generated and ground truth joint positions over the timesteps (N) and the number of test samples (S).
\begin{center}
APE$[j]=\dfrac{1}{SN} \sum_{s \in S} \sum_{n \in N} \| \widehat{H}_{s,n}[j] - H_{s,n}[j]\|_{2}$    
\end{center}

\noindent \textbf{Average Variational Error (AVE)} : For a joint j, AVE is calculated as the average of the L2 distances between the generated and ground truth variances.
\begin{center}
AVE$[j]=\dfrac{1}{S} \sum_{s \in S} \| \widehat{\sigma}_{s}[j] - \sigma_{s}[j]\|_{2}$    
\end{center}
where $\sigma[j]$ denotes the variance of the joint $j$,
\begin{center}
$\sigma[j]=\dfrac{1}{N-1} \sum_{n \in N} (\widetilde{H}[j] - H_{n}[j])^{2}$    
\end{center}
$\widetilde{H}[j]$ is calculated as the mean of the joint $j$ over $N$ timesteps.

\noindent We calculate four variants of errors for both APE and AVE,
\begin{itemize}
    \item \textit{root joint} error is calculated on the root joint using all the 3 coordinates $X, Y$ and $Z$.
    \item \textit{global traj} error is calculated on the root joint using only the 2 coordinates $X$ and  $Y$.
    \item \textit{mean local} error is calculated as the average of all the joint errors in the local coordinate system with respect to the root joint.
    \item \textit{mean global} error is calculated as the average of all the joint errors in the global coordinate system.
\end{itemize}

\section{Locomotion and root movement} 
Fig.~\ref{fig:locomotion} shows the diverse actions \textbf{involving locomotion} generated using Action-GPT. Our prompt strategy provides detailed descriptions which include the direction of the locomotion along with the coordination among relevant body parts. We further verify this by quantitatively evaluating using Average Positional Error and Average Variance Error of the global trajectory metric -- see Table.~\ref{tab:comparisions}. We observe that the Action-GPT variants are better aligned with root trajectory than the baseline.

\section{Current limitations}
\begin{itemize}
    \item \textit{Finger motion:} Current frameworks use SMPL to represent the pose. SMPL does not contain detailed finger joints. Therefore, GPT-generated descriptions for actions requiring detailed finger motion such as rock-paper-scissors, pointing fingers cannot be generated satisfactorily by the current framework.
    \item \textit{Complex Actions:} Actions containing complex and vigorous body movements such as yoga and dance poses cannot be generated by our current framework.
    \item \textit{Long duration action sequences:} Due to the limited duration of training data action sequences ($<10$ secs), our method cannot generate long sequences. 
\end{itemize} 

\begin{table}[!ht]
\centering
\resizebox{\linewidth}{!}
{
\begin{tabular}{c|l|c|c}
\toprule
Model & Method  & Avg. Training Time (secs) & Avg. Inference Time (secs)\\
\midrule
\multirow[c]{2}{*}[-0.5em]{\rule{0pt}{2ex} MotionCLIP}
& Default & 585 - 598 & 0.32 - 0.34 \\
& Action-GPT & 700 - 712 & 0.53 - 0.62  \\
\midrule
\multirow[c]{2}{*}[-0.5em]{\rule{0pt}{2ex} 
TEMOS}
& Default & 225 - 230 & 0.8 - 0.96 \\
& Action-GPT & 255 - 260 & 1.44 - 1.76   \\
\midrule
\multirow[c]{2}{*}[-0.5em]{\rule{0pt}{2ex} TEACH}
& Default & 234 - 240 & 1.6 - 2.1  \\
& Action-GPT & 315 - 320 & 3.4 - 3.8  \\
\bottomrule
\end{tabular}
}
\captionof{table}{Computation costs of baselines and their Action-GPT ($k=4$) variants.} 
\label{tab:computation_cost}
\end{table}

\section{Computation cost analysis}
There will be no change in the number of parameters of the Action-GPT variants when compared to the baseline models as we are using the frozen pre-trained text embedding models. Although the computation time of the models both during training and inference is higher for the Action-GPT ($k=4$) variants when compared with the original baselines as shown in Table.~\ref{tab:computation_cost}. For training time, we take the mean of the time taken for hundred epochs. For inference time, we take a batch-size of $16$ and average over $10$ repetitions. The increase in the computation time of the Action-GPT variants is because of the usage of $k$ GPT-3 text descriptions, each containing around $128$ words, whereas, in the baseline models, a single action phrase is used, which contains around $5-8$ words.
\end{document}